\documentclass[letterpaper, 10 pt, journal, twoside]{IEEEtran}
\usepackage{times,subfigure}
\usepackage{amsmath,amsfonts}
\usepackage{algorithmic}
\usepackage{mathtools}
\usepackage{algorithm}
\usepackage{array}
\usepackage{textcomp}
\usepackage{stfloats}
\usepackage{url}
\usepackage{verbatim}
\usepackage{graphicx}
\usepackage{cite}

\ifCLASSINFOpdf
\else
\fi

\begin{document}
\title{Diffusion Policies for Out-of-Distribution Generalization in Offline Reinforcement Learning}

% \author{Michael~Shell,~\IEEEmembership{Member,~IEEE,}
%         John~Doe,~\IEEEmembership{Fellow,~OSA,}
%         and~Jane~Doe,~\IEEEmembership{Life~Fellow,~IEEE}% <-this % stops a space
% \thanks{M. Shell was with the Department
% of Electrical and Computer Engineering, Georgia Institute of Technology, Atlanta,
% GA, 30332 USA e-mail: (see http://www.michaelshell.org/contact.html).}% <-this % stops a space
% \thanks{J. Doe and J. Doe are with Anonymous University.}% <-this % stops a space
% \thanks{Manuscript received April 19, 2005; revised August 26, 2015.}}
\author{Suzan Ece Ada$^{1}$, Erhan Oztop$^{2}$,~\IEEEmembership{Member,~IEEE,} and Emre Ugur$^{1}$%
\thanks{Manuscript received: September 3, 2023; Revised: December 16, 2023; Accepted: January 16, 2024.}%Use only for final RAL version
\thanks{This paper was recommended for publication by Editor Aleksandra Faust upon evaluation of the Associate Editor and Reviewers' comments.
This work was supported by Grant-in-Aid for Scientific Research – the project with number 22H03670, the project JPNP16007 commissioned by the New Energy and Industrial Technology Development Organization (NEDO), the Scientific and Technological Research Council of Turkey (TUBITAK, 118E923), and INVERSE project (101136067) funded by the European Union.} %Use only for final RAL version
\thanks{$^{1}$Suzan Ece Ada and Emre Ugur are with Department of Computer Engineering, Bogazici University, Istanbul, Turkey
        {\tt\footnotesize \{ece.ada,emre.ugur\}@bogazici.edu.tr}}%
\thanks{$^{2}$Erhan Oztop is with SISReC, OTRI, Osaka University, Japan, and Department of Computer Science, Ozyegin University, Istanbul, Turkey
        {\tt\footnotesize erhan.oztop@otri.osaka-u.ac.jp}}%
%\thanks{Digital Object Identifier (DOI): see top of this page.}
}

\markboth{IEEE Robotics and Automation Letters. Preprint Version. Accepted JANUARY, 2024}
{Ada \MakeLowercase{\textit{et al.}}: Diffusion Policies for Out-of-Distribution Generalization in Offline Reinforcement Learning} 




\maketitle

\begin{abstract}
Offline Reinforcement Learning (RL) methods leverage previous experiences to learn better policies than the behavior policy used for data collection.  However, they face challenges handling distribution shifts due to the lack of online interaction during training. To this end, we propose a novel method named State Reconstruction for Diffusion Policies (SRDP) that incorporates state reconstruction feature learning in the recent class of diffusion policies to address the problem of out-of-distribution (OOD) generalization. Our method promotes learning of generalizable state representation to alleviate the distribution shift caused by OOD states. To illustrate the OOD generalization and faster convergence of SRDP, we design a novel 2D Multimodal Contextual Bandit environment and realize it on a 6-DoF real-world UR10 robot, as well as in simulation, and compare its performance with prior algorithms. In particular, we show the importance of the proposed state reconstruction via ablation studies. In addition, we assess the performance of our model on standard continuous control benchmarks (D4RL), namely the navigation of an 8-DoF ant and forward locomotion of half-cheetah, hopper, and walker2d, achieving state-of-the-art results. Finally, we demonstrate that our method can achieve 167\% improvement over the competing baseline on a sparse continuous control navigation task where various regions of the state space are removed from the offline RL dataset, including the region encapsulating the goal.
\end{abstract}

% \begin{IEEEkeywords}
% IEEE, IEEEtran, journal, \LaTeX, paper, template.
% \end{IEEEkeywords}

% \fi
\IEEEpeerreviewmaketitle
\begingroup
\renewcommand\thefootnote{}\footnote{%
© 2024 IEEE. Personal use of this material is permitted.  Permission from IEEE must be obtained for all other uses, in any current or future media, including reprinting/republishing this material for advertising or promotional purposes, creating new collective works, for resale or redistribution to servers or lists, or reuse of any copyrighted component of this work in other works. DOI: 10.1109/LRA.2024.3363530
}%
\addtocounter{footnote}{-1}
\endgroup
\vspace*{-4mm}

\section{Introduction}
% for IEEE journal papers produced under \LaTeX\ using
% IEEEtran.cls version 1.8b and later.
\IEEEPARstart{L}{EVERAGING} large datasets and generalizing to unforeseen situations are critical components of intelligent systems. Offline Reinforcement Learning (RL) has garnered significant attention for learning from previously collected datasets without interacting with the real world \cite{fu2020d4rl}. Similarly, Out-of-Distribution (OOD) generalization is crucial for developing reliable systems adapting to unexpected conditions. While offline RL promises to find better policies than the behavior policy that generated the trajectories in the dataset without making assumptions about the agents' expertise, they struggle when faced with states not present in the training set. Hence, we aim to develop a generalizable generative offline RL model, State Reconstruction for Diffusion Policies (SRDP), to learn robust skills from offline datasets by extrapolating sequential decision-making to OOD states.

The main challenges of offline RL are the distribution shift and uncertainty estimation \cite{DBLP:journals/corr/abs-2005-01643}. Since state and action distributions in the offline dataset can differ from those encountered in the evaluation environment, dealing with OOD state and action samples is a prominent topic in offline RL \cite{li2021dealing}. Thus, we are interested in policy-regularized offline RL algorithms where the divergence from the behavior policy that collected the dataset is discouraged. 

There is a growing interest in applying diffusion models in robotics. Notably, Chi et al. \cite{chi_diffusion_visiomotor} have demonstrated the effectiveness of convolutional and transformer-based diffusion networks in representing multimodal action distributions with visuomotor policy learning for behavior cloning. Likewise, other behavior cloning \cite{florence2021implicit,zhao2023learning} methods have also learned multimodal expert behavior, though not specifically targeting offline RL and diffusion models. Despite these achievements, our current focus is on OOD generalization in offline RL without a visual component while believing in the potential applicability of our findings in visuomotor policy learning. Diffusion-QL \cite{wang2023diffusion} and Diffusers \cite{janner2022diffuser} are recent RL algorithms that utilize diffusion models \cite{DBLP:journals/corr/Sohl-DicksteinW15,NEURIPS2019_3001ef25,NEURIPS2020_4c5bcfec} by guiding the diffusion process toward regions that can yield a high reward. Diffusers \cite{janner2022diffuser} use diffusion models in the planning procedure to generate trajectories from the diffusion model, while Diffusion-QL \cite{wang2023diffusion} generates actions based on a state-conditional diffusion model and uses Q-function maximization with behavior cloning loss. Even though Diffusion-QL \cite{wang2023diffusion} can represent multimodal actions, it is often unstable in OOD state regions. 

Autoencoders can reconstruct a subset of OOD samples with low error via learning representations in the bottleneck layer \cite{denouden2018improving}. Denouden et al. \cite{denouden2018improving} highlight that OOD samples close to a linear or a nonlinear latent dimension manifold of the training data can have a low reconstruction loss. Hence, guidance from a reconstruction loss and similar architectural designs can benefit OOD generalization. Mutlu et al. \cite{MUTLU2020107320} use reconstruction loss to provide hints to the generator. Although we employ separate output heads to integrate the state reconstruction loss, we use a shared representation layer to generalize to OOD states close to the latent dimension manifold.

Our key contributions include SRDP, a new offline RL method that alleviates the distribution shifts incurred by OOD states using representation learning. SRDP is tailored to guide diffusion policies using a state reconstruction signal. Through extensive experiments, ablation studies, as well as real robot experiments, we demonstrate that incorporating state reconstruction signal at each diffusion timestep not only enhances the performance of foundational diffusion models in severe OOD settings where a large portion of data is missing during training but also in widely used offline RL tasks.

\section{Related Work}
\subsection{Offline Reinforcement Learning}
In \textbf{dynamic programming-based offline RL} methods, the Q-function is approximated using the offline RL dataset that comprises samples collected by the behavior policy. As the learned policy diverges from the behavior policy, Q-function estimates tend to exhibit overestimation in regions where uncertainty is high \cite{DBLP:journals/corr/abs-1906-00949}. Prior works impose a constraint on the statistical distance between the learned policy and the behavior policy in the policy optimization objective \cite{DBLP:journals/corr/abs-2005-01643} or the reward function \cite{DBLP:journals/corr/abs-1911-11361}. However, both methods require the computation of the behavior policy through behavior cloning and enforce a constraint on the learned policy. Nair et al. \cite{DBLP:journals/corr/abs-2006-09359} alleviated the need for computing the behavior policy by approximating the policy objective using an advantage-weighted maximum likelihood. Still, this method is prone to distribution shift as the policy can query the OOD actions during training. Recent work addressed this distribution shift by avoiding OOD actions in the Q-function estimation using an expected loss instead of the mean squared error loss \cite{DBLP:journals/corr/abs-2110-06169}. As an alternative, in Conservative Q-Learning (CQL) \cite{DBLP:journals/corr/abs-2006-04779}, the Q-function is approximated using a minimax objective where overestimated action values are minimized, and actions from the dataset are maximized. However, conservative estimation of the value function is prone to overfitting when data is scarce \cite{DBLP:journals/corr/abs-2005-01643}.

\textbf{Importance sampling-based offline RL} methods attempt to approximate the learned policy or expected return via off-policy RL techniques. However, off-policy RL allows environment interaction in the training loop, whereas offline RL learns from a static replay buffer constructed before training. The accuracy of the importance sampling estimator depends on the proximity of the learned policy to the policy used for data collection, the dimensionality of the state-action space, and the horizon of the task. Hence, these algorithms have limited applicability to real tasks and impose constraints on our objective of obtaining the best policy supported by the data. Prior work on importance sampling-based offline RL addresses the bias-variance trade-off. The marginal importance ratio \cite{pmlr-v119-zhang20r} can be used via dynamic programming to reduce bias. Doubly robust estimator \cite{DBLP:journals/corr/JiangL15} uses recursive regression-based value evaluation to reduce variance. However, these methods are still prone to distribution shifts, as they exploit OOD regions in the policy improvement step. Hence, this work uses an auxiliary state reconstruction signal in the policy improvement step that encourages learning more generalizable state features.

\textbf{Model-based offline RL} methods focus on deriving the environment model by estimating the transition function, unlike model-free methods. Although, theoretically, their advantage over model-free methods has not been proven \cite{DBLP:journals/corr/abs-2005-01643}, they offer sample efficiency and quick adaptation. In the model-based offline RL setting, the model cannot correct OOD states in addition to the OOD actions. To avoid OOD states and actions, a scaled uncertainty function penalizes the reward function over state action pairs \cite{DBLP:journals/corr/abs-2005-13239}. However, estimating an uncertainty function that accurately quantifies uncertainty regions over the state action space is a challenging open problem \cite{DBLP:journals/corr/abs-2005-01643}. Conservative Model-Based-RL \cite{DBLP:journals/corr/abs-2102-08363}, on the other hand, follows a similar approach used in CQL \cite{DBLP:journals/corr/abs-2006-04779} and penalizes overestimated Q-values of state action tuples sampled from the model distribution. Most model-based RL algorithms are myopic and fail to make accurate predictions in tasks where the dimensions of the state action space are high \cite{DBLP:journals/corr/abs-2005-01643}. Recent works have viewed the problem through the lens of sequence modeling and used high-capacity transformers \cite{DBLP:journals/corr/abs-2106-02039}, \cite{DBLP:journals/corr/abs-2106-01345}. However, these architectures are computationally expensive to train as they need careful tuning of hyperparameters. Validation and hyperparameter optimization in offline RL remains an open area of research.

\subsection{Diffusion Models}
Diffusion models are probabilistic generative models used in computer vision \cite{DBLP:journals/corr/Sohl-DicksteinW15,DBLP:journals/corr/abs-2301-08072,DBLP:journals/corr/abs-2301-09430,DBLP:journals/corr/abs-2301-10227}, natural language processing \cite{DBLP:journals/corr/abs-2209-12152,DBLP:journals/corr/abs-2211-13095}, and more recently, RL \cite{janner2022diffuser,wang2023diffusion}. Diffusion probabilistic models (DPM) \cite{DBLP:journals/corr/Sohl-DicksteinW15} formulate a forward diffusion process by adding a small amount of Gaussian noise to the data samples. By learning the reverse diffusion process, diffusion models learn to generate samples from Gaussian noise. Denoising diffusion probabilistic models (DDPMs) \cite{NEURIPS2020_4c5bcfec} explore DPM's relation to denoising score matching with annealed Langevin dynamics in image synthesis tasks. On the other hand, score-based generative models (SGMs) use a Noise Conditional-Score Network to learn scores at different levels of noise after perturbing the data for training stabilization \cite{https://doi.org/10.48550/arxiv.2209.00796}. Although our approach is developed for the offline RL framework, it can be applied to conditional diffusion models that use classifier guidance. In our case, however, the guidance comes from the states.

\section{Background}
\subsection{Offline Reinforcement Learning}
A Markov Decision Process (MDP) tuple is defined by $( \mathcal { S }, \mathcal{ A },\mathcal{ P}, r, \rho_0,\gamma)$ where $\mathcal{ S } $ is the state space of state   $\boldsymbol{s} \in \mathcal{ S } $, $\mathcal{ A } $ is the action space of action $\boldsymbol{a} \in \mathcal{ A } $, $\mathcal{P}{ (\boldsymbol{s'}|\boldsymbol{s}, \boldsymbol{a})}: \mathcal { S } \times \mathcal { A } \times \mathcal { S } \rightarrow [0,1] $ is the conditional probability distribution expressing dynamics, $\boldsymbol{s'}$ is the next state, $ r $ is the reward function, $\rho_0$ is the initial state distribution and $ \gamma \in ( 0,1 ] $ is the discount factor \cite{sutton2018reinforcement}. The objective of an RL agent is to learn a policy $\pi_\theta(\boldsymbol{a}|\boldsymbol{s})$, parameterized by $\theta$, that maximizes the expected cumulative discounted reward denoted by $\mathbb{E}_{\pi}\left[\sum_{t} \gamma^t r\left(\boldsymbol{s}_t, \boldsymbol{a}_t\right)\right]$. The state-action value function, Q-function, $Q^\pi(\boldsymbol{s}, \boldsymbol{a})$ is defined as the expected cumulative discounted reward gained for taking an action $\boldsymbol{a}$ at a state $\boldsymbol{s}$ then following $\pi$. In offline RL, the agent is tasked with learning the best policy supported by the dataset of MDP tuples denoted by $\mathcal{D}=\{(\boldsymbol{s}_i,\boldsymbol{a}_i,\boldsymbol{s}'_i,r_i)\}$. The dataset is constructed from the rollouts obtained by a behavior policy $\pi_\beta( \boldsymbol{a}|\boldsymbol{s})$ \cite{DBLP:journals/corr/abs-2005-01643}.  

\subsection{Diffusion Models}
Diffusion probabilistic models \cite{DBLP:journals/corr/Sohl-DicksteinW15,NEURIPS2019_3001ef25,NEURIPS2020_4c5bcfec}, commonly called diffusion models for brevity, are a class of probabilistic generative models that seek to generate new samples by learning the underlying probability distribution of the data. The forward diffusion process follows a Markov chain to slowly destroy the structure of an original data sample, $x_0$, by adding noise to obtain a sequence of noisy samples $x_1...x_T$. Here, the Gaussian noise added to the data depends on a variance schedule $\{\beta_t \in (0,1)\}^T_{t=1}$, where $\beta_t$ is the diffusion rate at timestep t. Since the sequences of noisy samples are available during training, we can train a neural network to predict the noise $\boldsymbol{\epsilon}_t$ added to the data at a given timestep. At a high level, we can generate samples from Gaussian noise through an iterative denoising procedure in the reverse diffusion process by using the predicted noise added at each timestep.

In diffusion models, we can directly sample the noisy image at any time t. By replacing the diffusion rate $\beta_t$ with $\alpha_t=1-\beta_t$, we obtain the distribution $q\left(\mathbf{x}_t \mid \mathbf{x}_0\right)=\mathcal{N}\left(\mathbf{x}_t ; \sqrt{\bar{\alpha}_t} \mathbf{x}_0,\left(1-\bar{\alpha}_t\right) \mathbf{I}\right)$ using recursion and reparametrization technique where $\bar{\alpha}_t=\prod_{i=1}^t \alpha_i$. To estimate the reverse process, we learn the parameters of $p_\theta\left(\mathbf{x}_{t-1} \mid \mathbf{x}_t\right)=\mathcal{N}\left(\mathbf{x}_{t-1} ; \boldsymbol{\mu}_\theta\left(\mathbf{x}_t, t\right), \mathbf{\Sigma}_\theta\left(\mathbf{x}_t, t\right)\right)$. The joint distribution of the reverse diffusion is denoted by 
$p_\theta\left(\mathbf{x}_{0: T}\right)=p\left(\mathbf{x}_T\right) \prod_{t=1}^T p_\theta\left(\mathbf{x}_{t-1} \mid \mathbf{x}_t\right) \quad$ where $\mathbf{x}_T$ is the isotropic Gaussian distribution. Ho et al. \cite{NEURIPS2020_4c5bcfec}, formulated the simplified loss function in DDPMs for diffusion timestep $t$ 
$$
L_t=\mathbb{E}_{t, \mathbf{x}_0, \boldsymbol{\epsilon}_t}\left[\left\|\epsilon_t-\boldsymbol{\epsilon}_\theta\left(\sqrt{\bar{\alpha}_t} \mathbf{x}_0+\sqrt{1-\bar{\alpha}_t} \boldsymbol{\epsilon}_t, t\right)\right\|^2\right]
$$
where $\boldsymbol{\epsilon}_\theta\left(\sqrt{\bar{\alpha}_t} \mathbf{x}_0+\sqrt{1-\bar{\alpha}_t} \boldsymbol{\epsilon}_t, t\right)$ is the noise predicted by the neural network and $\boldsymbol{\epsilon}_t$ is the true noise used in the forward process.

\section{Method}
To address the OOD states in Offline RL, we propose incorporating an auxiliary state reconstruction loss in diffusion policies. We first describe the details of Diffusion Q-Learning (Diffusion-QL)\cite{wang2023diffusion} and 
its implementation. Subsequently, we derive a robust diffusion policy that leverages representation learning. Finally, we discuss our approach using a 2D multimodal contextual bandit environment.

\subsection{Diffusion Q-Learning}
Diffusion-QL uses a conditional diffusion model to generate actions conditioned on states \cite{wang2023diffusion}. The conditional reverse diffusion chain conditioned on state $s$ is defined by $\pi_\theta\left(\boldsymbol{a}_{0: T} \mid \boldsymbol{s}\right)$ where T denotes the diffusion timestep. The action obtained by the reverse diffusion process is then used in Q-learning and policy learning in an iterative procedure. Initially, a minibatch of MDP tuples $\{(s, a,r,s^{\prime})\}$ are sampled from the offline RL dataset. The next action $a^{\prime}$ is generated by the target diffusion policy $\pi_{\theta^{\prime}}$ conditioned on the next state $s^{\prime}$. Using double Q-learning trick by \cite{van2016deep} and Bellman operator minimization by \cite{Fujimoto2018OffPolicyDR,DBLP:journals/corr/LillicrapHPHETS15}, Diffusion-QL, minimizes
\begin{equation}
\scriptstyle{
\mathbb{E}_{\substack{\left(\boldsymbol{s}, \boldsymbol{a}, \boldsymbol{s}^{\prime}\right) \sim \mathcal{D} \\\boldsymbol{a}^{\prime}_0 \sim \pi_{\theta^{\prime}} }}\left[\left\|\left(r\left(\boldsymbol{s}, \boldsymbol{a}\right)+\gamma \min _{i=1,2} Q_{\phi_i^{\prime}}\left(\boldsymbol{s}^{\prime}, \boldsymbol{a}^{\prime}_0\right)\right)\\-Q_{\phi_i}\left(\boldsymbol{s}, \boldsymbol{a}\right)\right\|^2\right]
}
\label{eq:doubleqlearning}
\end{equation}
\noindent where $\mathcal{D}$ is the offline RL dataset, $Q_{\phi_i^{\prime}}$ are the target critic networks and $Q_{\phi_i}$ are the critic networks. Diffusion policies are optimized by policy improvement using Q-function and behavior cloning loss minimization. We update the critic networks similarly to evaluate the improvement from the state reconstruction loss.

\subsection{State Reconstruction for Diffusion Policies}
Diffusion policies learn to generate actions with the guidance of states. Since state information is available in the training and evaluation phase, the diffusion policy can be conditioned on the state to generate actions. The diffusion model learns to predict the noise $\epsilon_t$ added to the input at each diffusion iteration $\mathbf{t}$ following the simplified loss in DDPM. Diffusion policies extend this idea to the RL framework by concatenating the noisy input (noisy action) vector and the embedded diffusion timestep with the state vector during training. In contrast, using an auxiliary head, SRDP learns to extract the state representation in a shared representation layer. In brief, SRDP operates as follows. The shared SRDP feature extraction module $f_\phi$ maps the time embedding, state, and noisy action to a latent representation $\boldsymbol{z}$. This representation is then directed into distinct task-specific heads: the diffusion head $f_\theta$ predicts the added noise at that diffusion timestep for the given noisy action and state, and the auxiliary head $f_\psi$ reconstructs the input state. Importantly, the state representation signal is deeply embedded for all diffusion timesteps sampled during training through the state reconstruction loss.

We use a shared fully connected module $f_\phi$ to map the noisy action $\sqrt{\bar{\alpha}_t} \boldsymbol{a}+\sqrt{1-\bar{\alpha}_t} \boldsymbol{\epsilon}$, time-embedding and the state $\boldsymbol{s}$ into a shared representation 
\begin{equation}
\boldsymbol{z}=f_\phi\left(\sqrt{\bar{\alpha}_t} \boldsymbol{a}+\sqrt{1-\bar{\alpha}_t} \boldsymbol{\epsilon}, \boldsymbol{s}, t\right).
\end{equation}
The diffusion head $\boldsymbol{f}_\theta$ uses this representation to predict the noise added at timestep $t$, while state head $\boldsymbol{f}_\psi$ learns to reconstruct the state. Thus, diffusion policy loss $\mathcal{L}_{\boldsymbol{DP}}$ is defined by 
\begin{equation}
 \scriptstyle{
\mathbb{E}_{\substack{t \sim \textit{U}(\left\{\textit{1,...,T}\right\}) \\ \boldsymbol{\epsilon} \sim \mathcal{N}(\mathbf{0}, \boldsymbol{I})}}\left[\left\|\boldsymbol{\epsilon}-f_\theta\left(f_\phi\left(\sqrt{\bar{\alpha}_t} \boldsymbol{a}+\sqrt{1-\bar{\alpha}_t} \boldsymbol{\epsilon}, \boldsymbol{s}, t\right)\right)\right\|^2\right]}
\end{equation}
where diffusion timesteps are denoted by $t$. The diffusion timesteps are sampled from the uniform distribution $U$ over the set $\left\{\textit{1,...,T}\right\}$. State reconstruction guidance is also propagated through the network during policy learning. The state reconstruction loss $\mathcal{L}_{\boldsymbol{R}}$ minimizes 
\begin{equation}
 \scriptstyle{
\mathbb{E}_{\substack{t \sim \textit{U}(\left\{\textit{1,...,T}\right\}) \\ \boldsymbol{\epsilon} \sim \mathcal{N}(\mathbf{0}, \boldsymbol{I})}}\left[\left\|\boldsymbol{s}-f_\psi\left(f_\phi\left(\sqrt{\bar{\alpha}_t} \boldsymbol{a}+\sqrt{1-\bar{\alpha}_t} \boldsymbol{\epsilon}, \boldsymbol{s}, t\right)\right)\right\|^2\right]}
\end{equation}
where $f_\psi$ takes the shared representation as input to predict the state. Behavior cloning loss $\mathcal{L}_{\boldsymbol{BC}}$ in SRDP, jointly learn the contextual representation of the state and the noise added to the action by minimizing 
\begin{equation}
\mathcal{L}_{\boldsymbol{BC}}=\mathcal{L}_{\boldsymbol{DP}}+\lambda\mathcal{L}_{\boldsymbol{R}}
\label{eq:lbcloss}
\end{equation}
where $\lambda$ is a hyperparameter that controls the weight of the state reconstruction loss. A key point to note is that the state reconstruction loss in this step is propagated through the network for each randomly selected diffusion timestep.

At each training iteration, we first sample a mini-batch of transitions $\{(s,a,r,s^{\prime})\}$ from the offline RL dataset. Then, we sample a mini-batch of next actions $\{(a_{0}^{\prime}\}$ from the target diffusion policy using $\{(s^{\prime})\}$ from the offline RL dataset\cite{wang2023diffusion}. The mini-batch of transitions and the sampled $\{(a_{0}^{\prime}\}$ are used to update the critic network using Eq. \ref{eq:doubleqlearning} following the implementation in \cite{wang2023diffusion,dpcode}. After the critic update, we sample $\{(a_{0})\}$ from our policy $\pi_\theta$ through an iterative denoising procedure. To guide the action generation procedure to high reward regions, we subtract the scaled \cite{fujimoto2021a} expectation of Q-function, $\frac{\eta\mathbb{E}_{\boldsymbol{s} \sim \mathcal{D}, \boldsymbol{a}_0 \sim \pi_\theta}\left[Q_\phi\left(\boldsymbol{s}, \boldsymbol{a}_0\right)\right] }{\left.\mathbb{E}_{(s, a) \sim \mathcal{D}}\left\|Q_\phi(\boldsymbol{s}, \boldsymbol{a})\right\|\right]}$, from SRDP $\mathcal{L}_{\boldsymbol{BC}}$ in Eq. \ref{eq:lbcloss} to obtain $\mathcal{L}_{\boldsymbol{SRDP}}$. Crucially at this minimization step, the state reconstruction loss is propagated through the network for each diffusion timestep, promoting learning more descriptive features from the OOD-state samples. Then, we update the target policy network and the target critic networks.

In the evaluation phase, we sample actions from our policy through an iterative denoising procedure. First, we sample $a_T\sim\mathcal{N}(\mathbf{0}, \boldsymbol{I})$ at diffusion timestep T. Then, we concatenate this sampled action $a_T$ with our current state from the environment and the embedding of the diffusion timestep to obtain the input vector for our diffusion policy. Subsequently, our diffusion model predicts the noise $\scriptstyle{\epsilon_\theta=f_\theta\left(f_\phi\left(\boldsymbol{a}_t, \boldsymbol{s}, t\right)\right)}$ added to the action at that diffusion timestep given the state while the reconstructed states from $f_\psi$ are not used. To obtain the action $a_0$ generated by our model, we run the reverse diffusion chain by computing $\scriptstyle{\boldsymbol{a_{t-1}}|\boldsymbol{a_t}=\frac{1}{\sqrt{\alpha_t}}\left(\boldsymbol{a}_t-\frac{1-{\alpha}_t}{\sqrt{1-\bar{\alpha}_t}} f_\theta\left(f_\phi\left(\boldsymbol{a}_t, \boldsymbol{s}, t\right)\right)\right)+\sqrt{\beta_t} \mathbf{\epsilon}}$. As suggested in \cite{NEURIPS2020_4c5bcfec} to enhance the sampling quality, we sample $\mathbf{\epsilon}\sim\mathcal{N}(\mathbf{0}, \mathbf{I})$ for $t= T,\ldots,2$ and assign $\mathbf{\epsilon}=0$ for $t=1$.

\subsection{Out-of-Distribution States in Offline RL}
We aim to learn a robust policy from offline RL datasets consisting of expert and novice demonstrations with different modalities. Therefore, we designed an environment named 2D multimodal contextual bandit to evaluate the agent's performance in OOD states. Previous work \cite{wang2023diffusion} used a 2D-bandit environment to illustrate the advantage of using a highly expressive model to represent policies. Here, we extend this environment to a contextual bandit setting to assess the trained policy in OOD states. The primary objective in this task is to learn multimodal expert behavior, particularly in challenging OOD states. Thus, Double Q-learning \cite{van2016deep} is not used in this context to isolate the regularization induced by SRDP behavior cloning loss.

We consider a two-dimensional continuous state and action space where the states and actions are characterized as real-valued x- and y-coordinates. To illustrate multimodal expert behaviors, the states in each pair of quadrants map to actions generated from a mixture of two Gaussian distributions. Given a state (x,y), the agent needs to generate an action $[a_{1}(x,y), a_{2}(x,y)]$ that has a high probability in the matching mixture of two Gaussians. Subsequently, we construct a training dataset by sampling states from two uniform distributions and actions from two Gaussian mixture models (GMM), where each Gaussian mixture comprises two Gaussian distributions.  
For a given state $s$, an action $a$ is sampled from a GMM with two Gaussians as $a \sim \pi \mathcal{N}(\boldsymbol{\mu}_1(s), \boldsymbol{\Sigma}) + \pi \mathcal{N}(\boldsymbol{\mu}_2(s), \boldsymbol{\Sigma})$ where $\boldsymbol{\Sigma}$ is the diagonal covariance matrix with $\sigma=[0.05,0.05]$, $\pi$ is the mixture weight 0.5, $\boldsymbol{\mu}_1$, and $\boldsymbol{\mu}_2$  are the mean vectors of Gaussian distributions. We use the following procedure to determine the values of the mean vectors $\boldsymbol{\mu}_1(s)$, and $\boldsymbol{\mu}_2(s)$
\begin{equation}
    \scriptstyle{\mu_1(s),\mu_2(s)} = 
\begin{dcases}
    \scriptstyle{(-0.8,-0.8), (0.8,0.8)} & \scriptstyle{\text{if } s \in [-1.0,0.0] \times[0.0,1.0]}\\  
    \scriptstyle{(-0.8,-0.8), (0.8,0.8)} & \scriptstyle{\text{if } s \in [0.0,1.0]\times[-1.0,0.0]}\\
    \scriptstyle{(-0.8,0.8), (0.8,-0.8)} & \scriptstyle{\text{if } s \in [-1.0,0.0]\times[-1.0,0.0]}\\ 
    \scriptstyle{(-0.8,0.8), (0.8,-0.8)} & \scriptstyle{\text{if } s \in [0.0,1.0]\times[0.0,1.0]}
\end{dcases}.
\label{eq:actiongen}
\end{equation}
where a state (s) is sampled from the uniform distributions ($s_{train} \sim \mathcal{U}([-0.05,0.05]\times[-0.05,0.05])$) and ($s_{train} \sim \mathcal{U}([-0.08,0.08]\times[-0.08,0.08])$) when constructing the training data. During testing, we sample states from subregions within the Euclidean plane larger than those in the training dataset. In particular, we sample states from the uniform distribution $\mathcal{U}([-1.0,1.0]\times[-1.0,1.0])$ to evaluate the policies in OOD states.

 \begin{figure}
  \vspace*{0.5mm}
        \subfigure[]{
		\includegraphics[width=0.165\columnwidth]{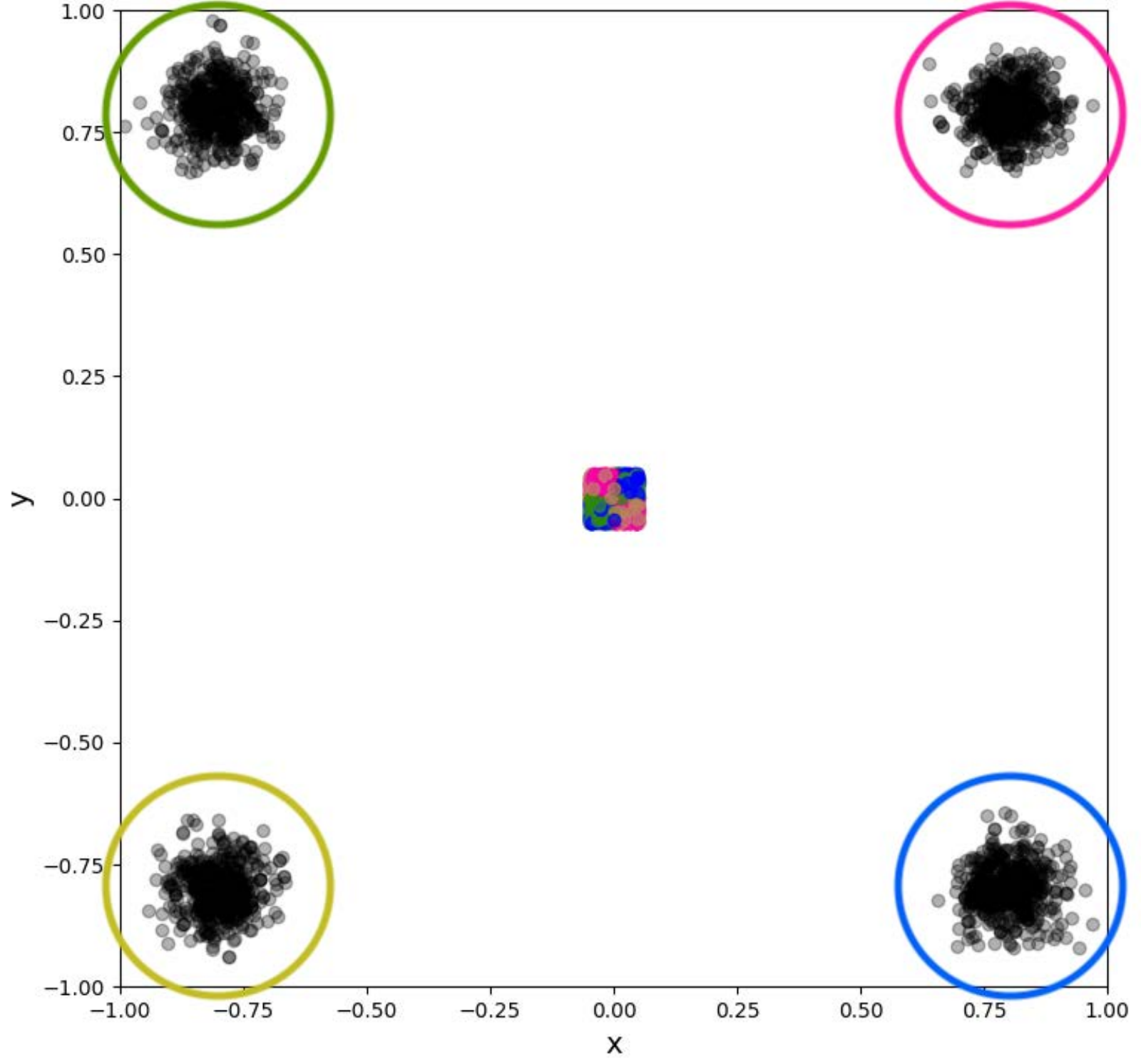}
	}
        \subfigure[]{
		\includegraphics[width=.165\columnwidth]{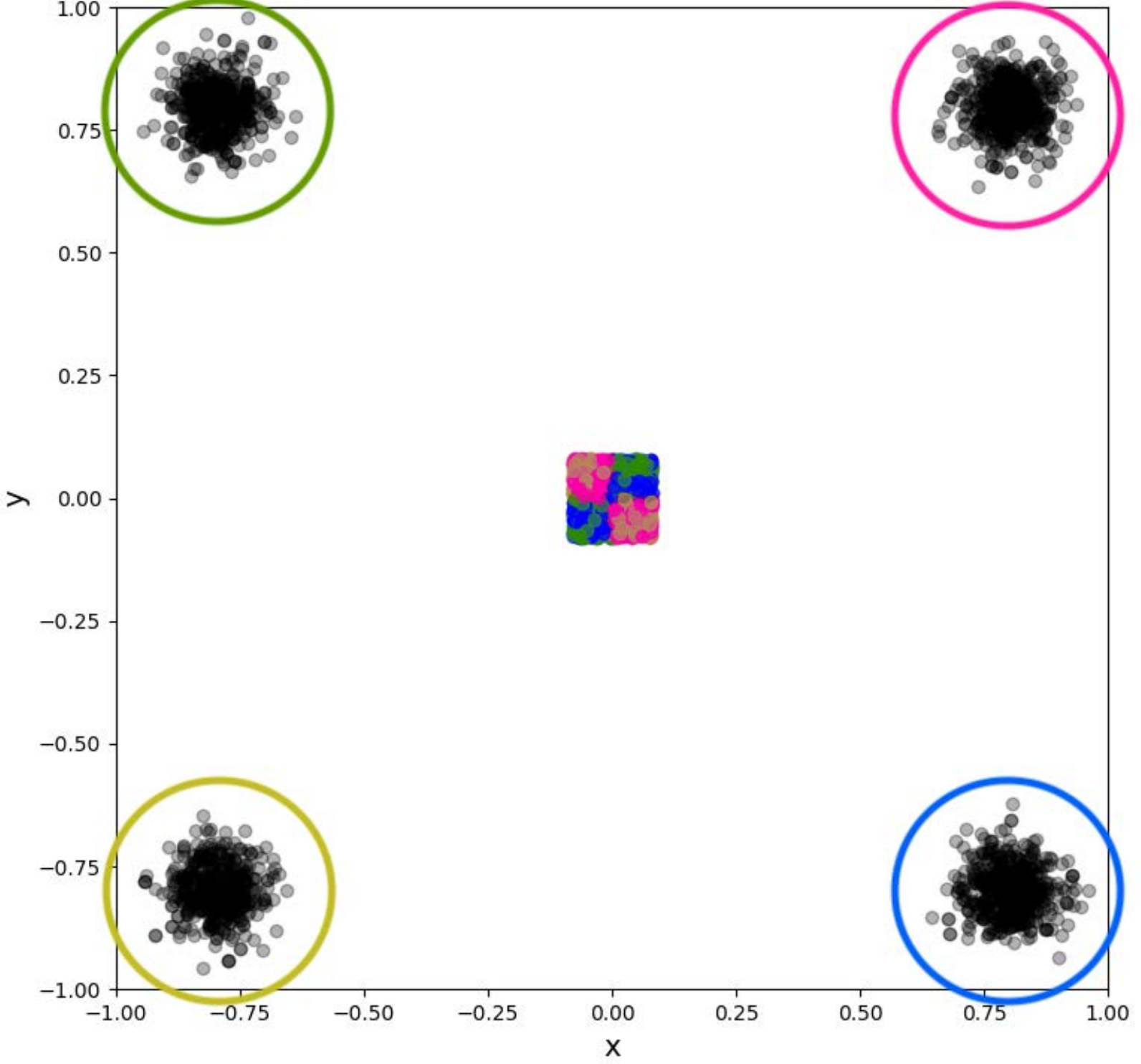}
            }
        \subfigure[]{
		\includegraphics[width=.16\columnwidth]{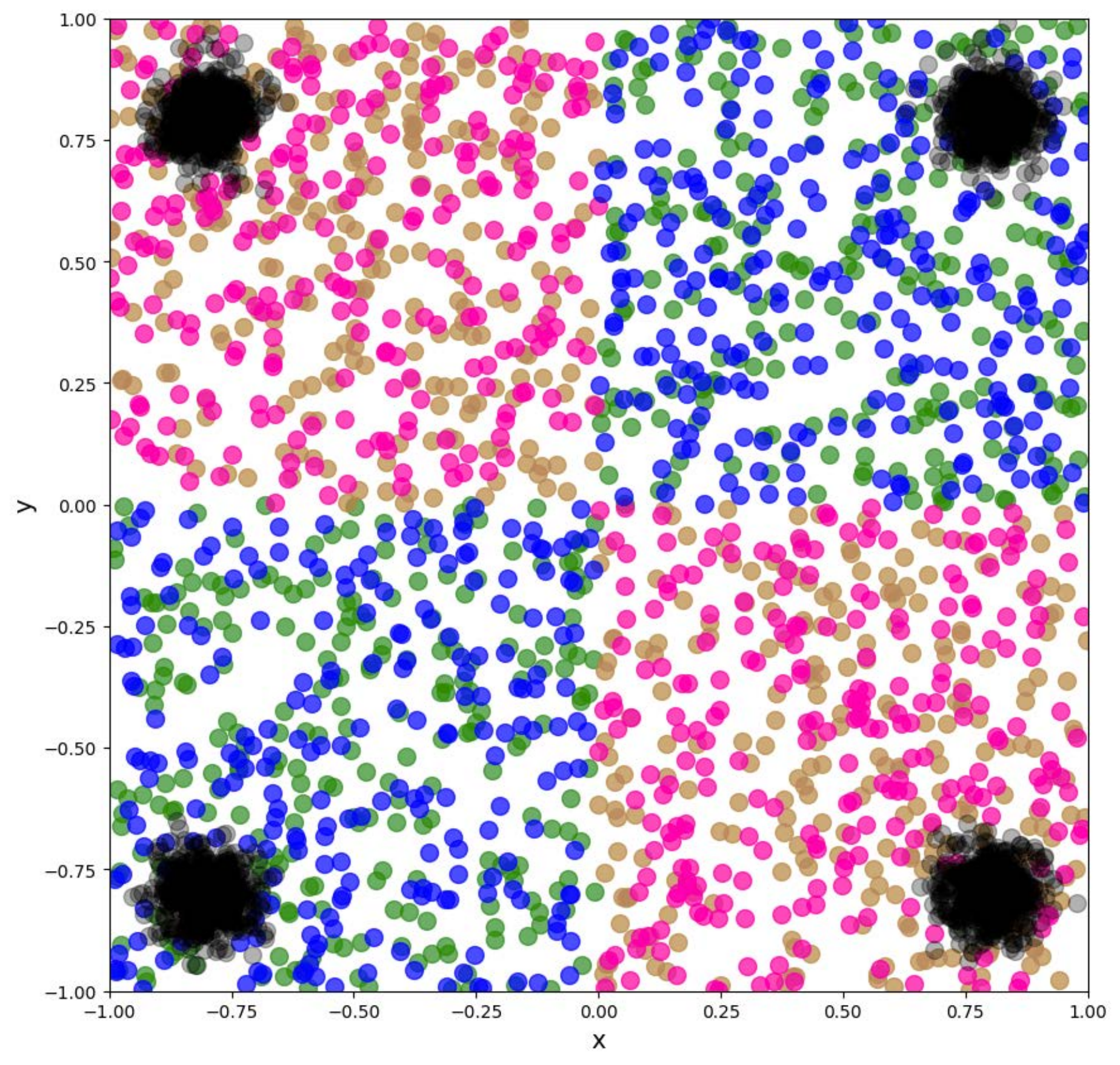}
            }
         \subfigure[]{
        \includegraphics[width=.165\columnwidth]{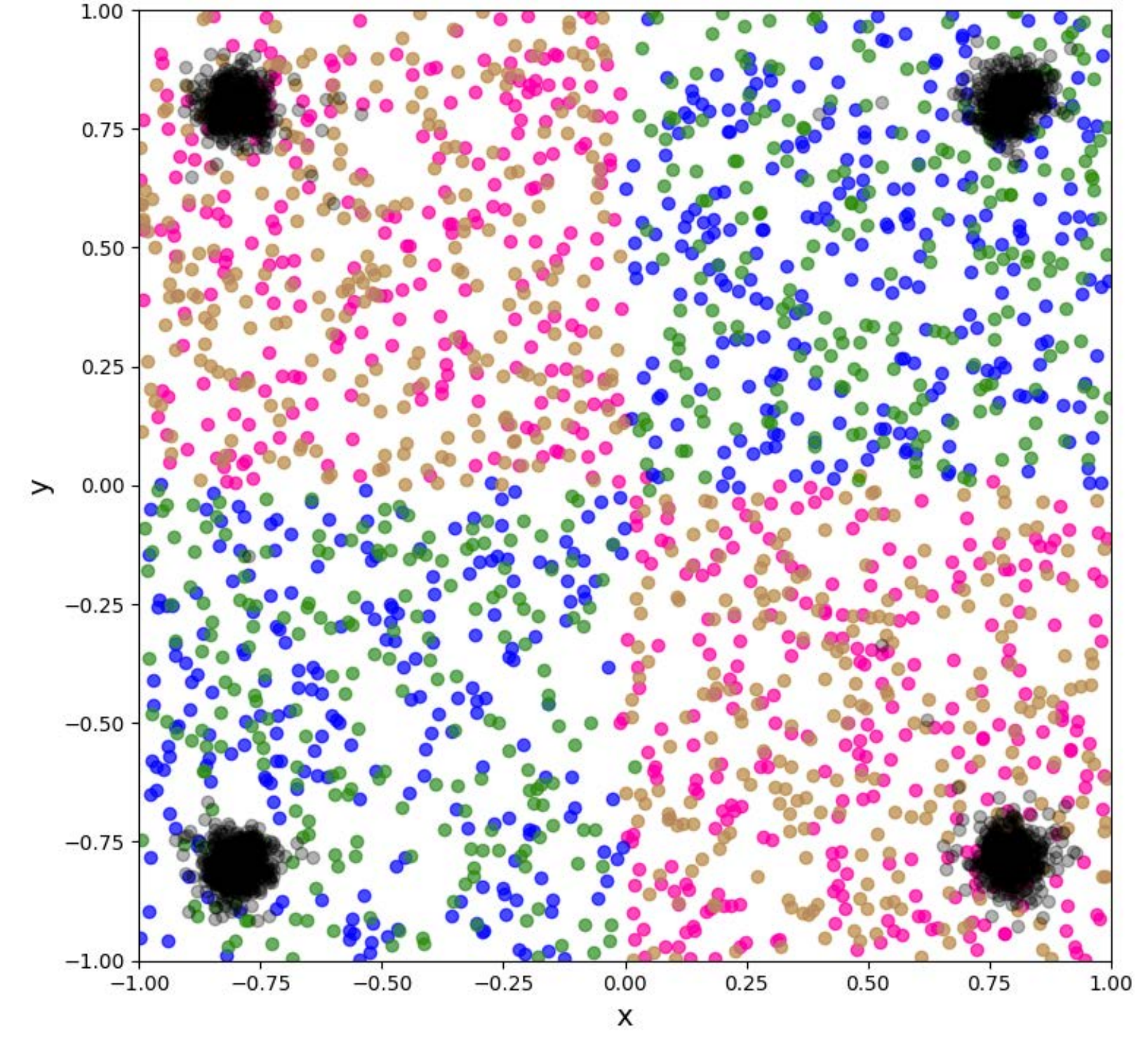}
            }
        \subfigure[]{
        \includegraphics[width=.165\columnwidth]{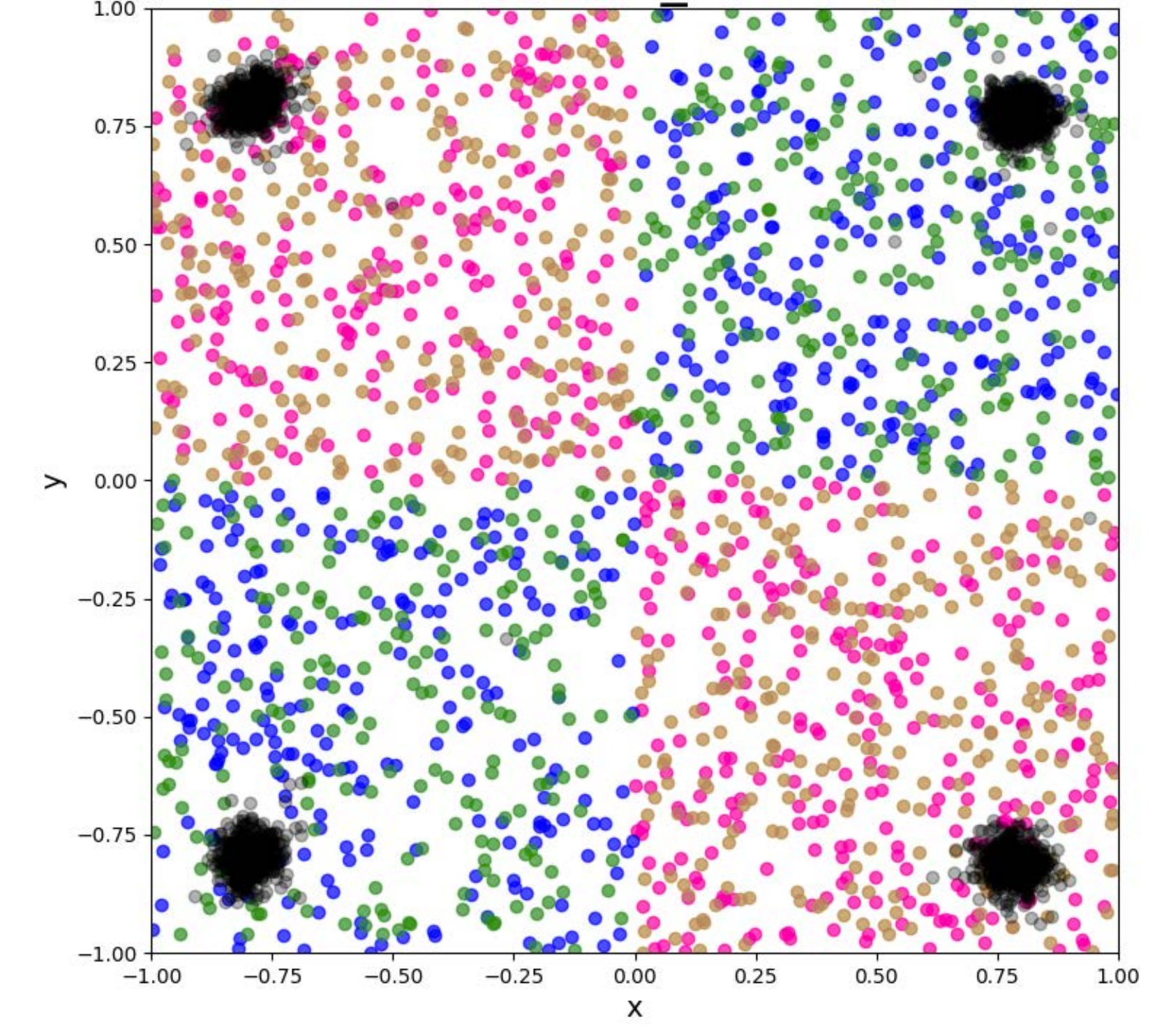}
            }
	\caption{The states mapped to actions (represented by black dots) in the first, second, third, and fourth quadrants are colored magenta, green, light brown, and blue, respectively. The training dataset (a) is constructed with $s_{train} \sim \mathcal{U}([-0.05,0.05]\times[-0.05,0.05])$ and used for experiments illustrated in Fig. \ref{fig:lim05hid16}. The training dataset in (b) is constructed from $s_{train} \sim \mathcal{U}([-0.08,0.08]\times[-0.08,0.08])$ and used for experiments illustrated in Fig. \ref{fig:srdpabl08} and Fig. \ref{fig:lim08hid16}. State samples for ground truth (c) are generated from the uniform distribution $\mathcal{U}([-1.0,1.0]\times[-1.0,1.0])$. Actions are generated from the GMM detailed in Eq. \ref{eq:actiongen}. Results for training with non-OOD data generated from the uniform distribution $\mathcal{U}([-1.0,1.0]\times[-1.0,1.0])$ are illustrated for BC-Diffusion (d) and SRDP(ours) (e). Chamfer distance between the ground truth contextual action distribution and the actions generated using the non-OOD dataset for BC-Diffusion is 0.047, and SRDP(ours) is $\mathbf{0.037}$.}
	\label{fig:offlinedataset}
\end{figure}

\section{Experiments}
This section empirically evaluates our proposed approach, i.e., state reconstruction guidance on the 2D-Multimodal Contextual Bandit environment, multimodal real-robot setup, sparse reward continuous control maze navigation dataset with missing data generated in \cite{mark2023offline} for offline RL pretraining, and D4RL benchmarks from \cite{fu2020d4rl}. 

\subsection{2D-Multimodal Contextual Bandit Experiments}
\begin{figure}[htbp]
	\centering
		\includegraphics[width=\columnwidth]{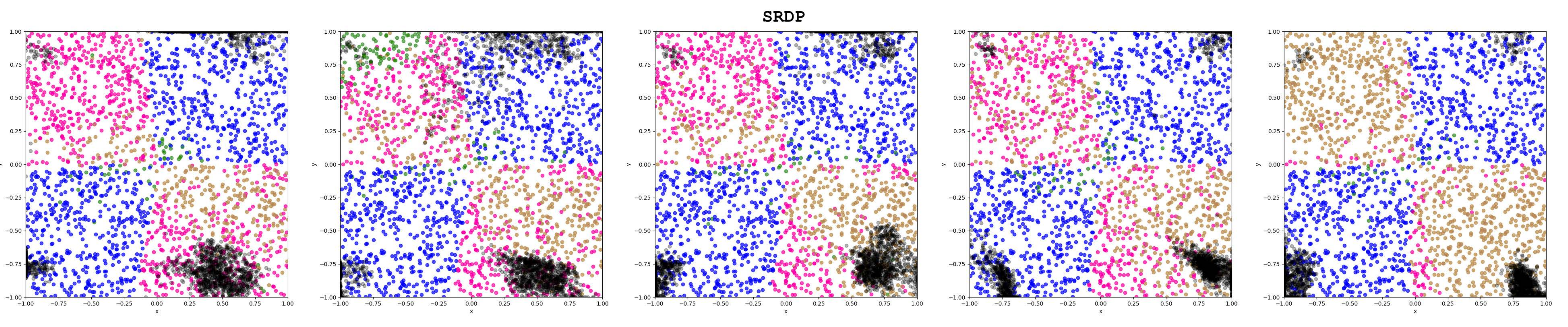}
		\includegraphics[width=\columnwidth]{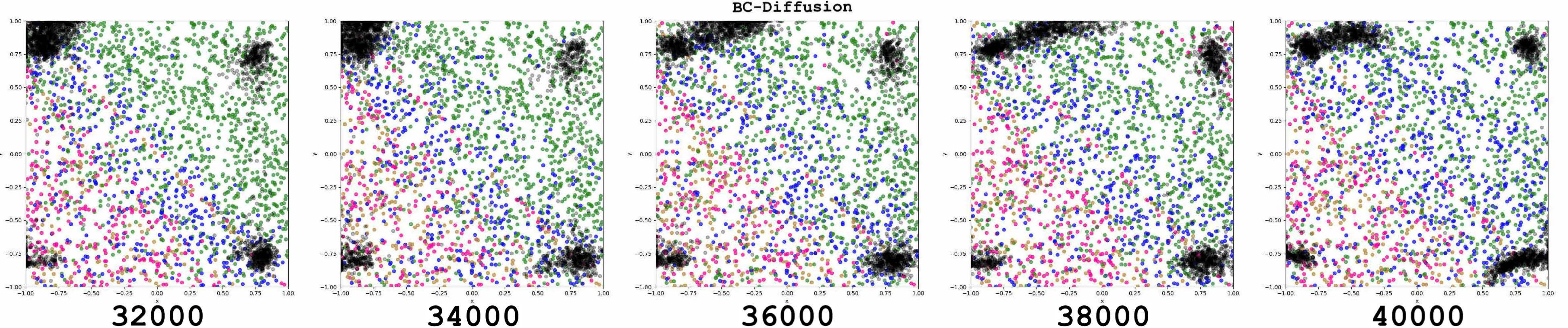}
	\caption{Top to bottom: SRDP $(\lambda=1.25)$ and BC-Diffusion \cite{wang2023diffusion}. Dataset for training is constructed from Fig. 1 (a) ($s_{train} \sim \mathcal{U}([-0.05,0.05]\times[-0.05,0.05])$) and ground truth is illustrated in Fig. \ref{fig:offlinedataset} (c). The number of training iterations in each column increases incrementally from left to right by 2000, starting from 32000, as the convergence for the more restricted dataset takes longer.}
	\label{fig:lim05hid16}
\end{figure}

The training dataset in 2D-Multimodal Contextual Bandit $D=\{(s_i,a_i)\}^n_{i=1}$, consists of $n=10000$ state-action tuples with horizon $H=1$. To examine the policy in OOD states, the states generated for training are sampled from the uniform distribution $s_{train} \sim \mathcal{U}([-0.05,0.05]\times[-0.05,0.05])$ in Fig. \ref{fig:offlinedataset}(a), and $s_{train} \sim \mathcal{U}([-0.08,0.08]\times[-0.08,0.08])$ in Fig. \ref{fig:offlinedataset}(b). The states used for the evaluation are sampled from the uniform distribution of $s_{test} \sim \mathcal{U}([-1,1]\times[-1,1])$. Ground truth for OOD state generalization corresponding to the training datasets in Fig. \ref{fig:offlinedataset}(a) and (b) is illustrated in Fig. \ref{fig:offlinedataset}(c) where the black points show the actions in the dataset following Eq. \ref{eq:actiongen}. Similarly, the black points in Fig. \ref{fig:lim05hid16}(a) and (b) show the actions generated by the SRDP and BC-Diffusion policies, respectively. If a state generates an action in the first, second, third, and fourth quadrants, the state is colored magenta, green, light brown, and blue, respectively.

Visually distinguishable quadrants with respective coloring in Fig. \ref{fig:lim05hid16}(a) (SRDP) indicate that the reverse diffusion process can generate actions accurately for OOD states. In contrast, Fig. \ref{fig:lim05hid16}(b) shows that BC-Diffusion memorizes actions without using state information. Subsequently, results in Fig. \ref{fig:lim05hid16} show that the representation learning with state reconstruction loss promotes finding expert skills in multimodal data, achieving faster convergence. Although a larger portion of the state space is not included in the training set distribution in Fig. \ref{fig:lim05hid16}, SRDP can generalize well to OOD states and learn at least one of the two expert behaviors. Conversely, BC-Diffusion finds the action distribution of the dataset, yet it cannot assign the correct action distribution for the given state. Accurate partitioning of the state space, in terms of the effect of the actions taken, is particularly important in real-world robotics tasks where naive memorization of the action distribution in the data without state dependence can have severe consequences. Although SRDP can represent a single mode in this challenging OOD task where only 0.25\% of the state space is present in the data, it can correctly learn an expert policy. All hidden layers have the same size of 16 for SRDP and BC-Diffusion.

\begin{figure}
 \vspace*{0.5mm}
  \centering
  \includegraphics[width=1\columnwidth]{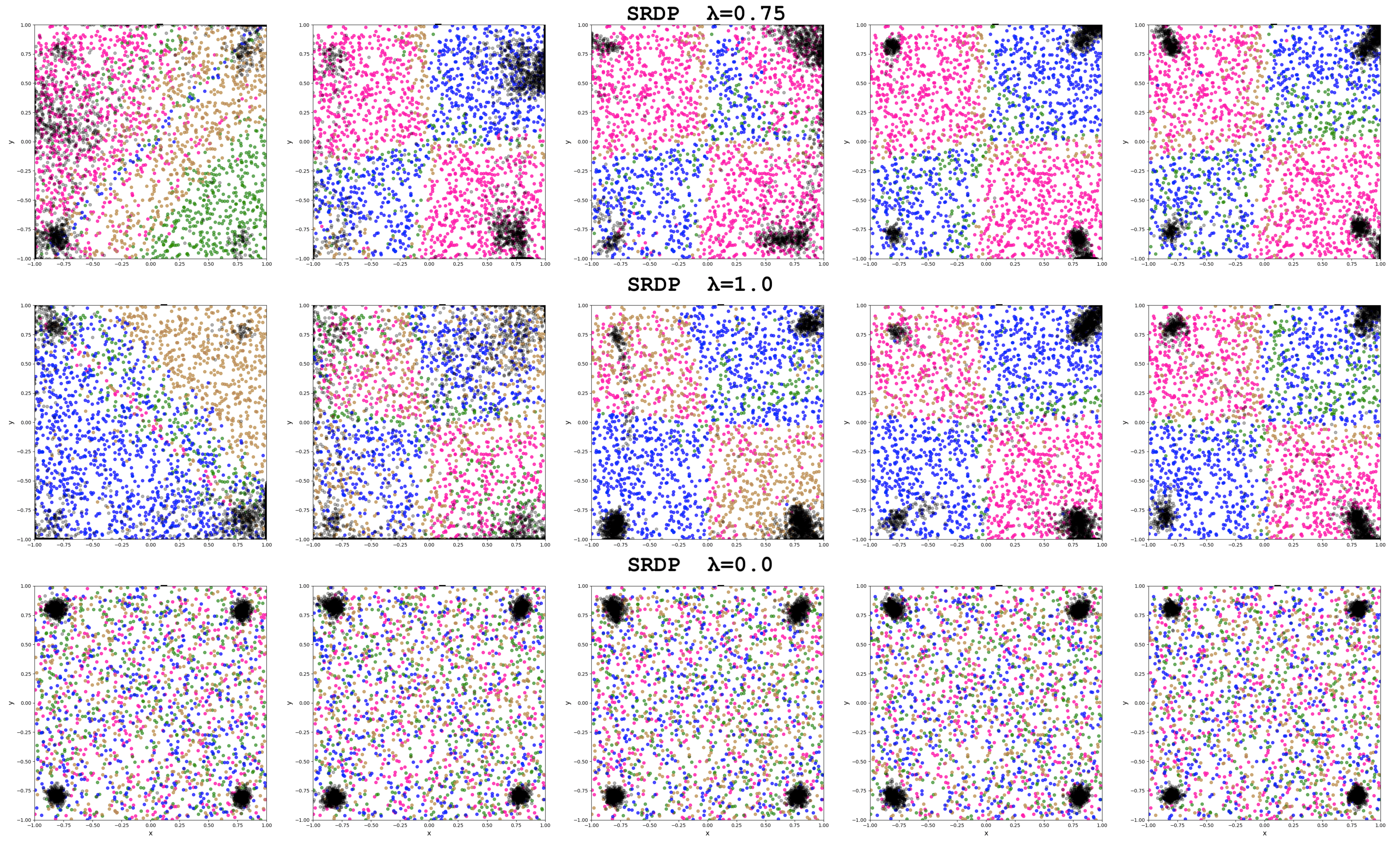}
   \includegraphics[width=1\columnwidth]{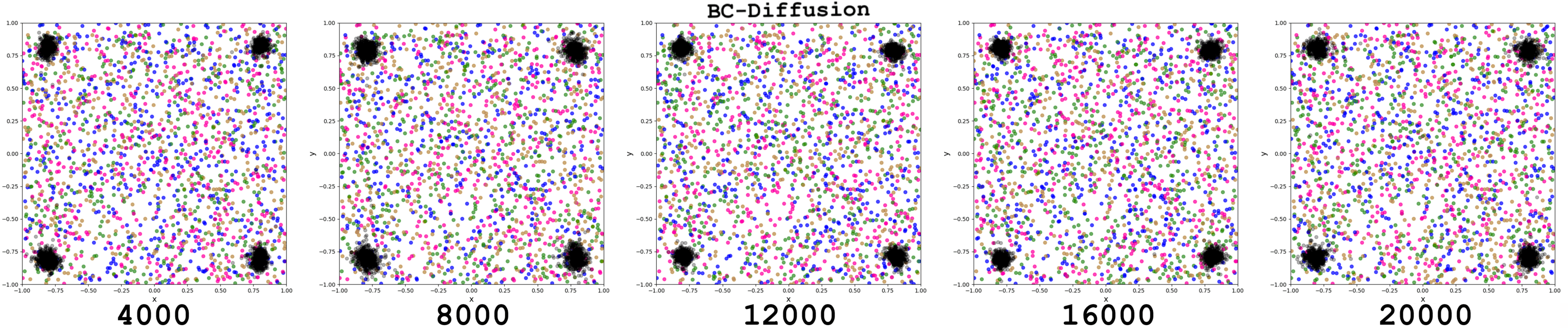}
 \caption{Top to bottom rows show the results from SRDP (proposed model) with the scaling parameter $\lambda=0.75$, $\lambda=1.0$, $\lambda=0.0$ (meaning no state reconstruction loss) and BC-Diffusion \cite{wang2023diffusion}. The training dataset is constructed from $s_{train} \sim \mathcal{U}([-0.08,0.08]\times[-0.08,0.08]$). The number of training iterations increases incrementally by 4000 from left to right.}
\label{fig:srdpabl08}
\end{figure}

In the next set of experiments, the training dataset includes $s_{train} \sim \mathcal{U}([-0.08,0.08]\times[-0.08,0.08]$) and $a_{train}$ generated by Eq. \ref{eq:actiongen}. Fig. \ref{fig:srdpabl08} and Fig. \ref{fig:lim08hid16} show that only SRDP can represent multimodal expert distributions and partition the state space into visible quadrants at earlier iterations, improving training stability and reducing computation costs. Consistent with the experiments in Fig. \ref{fig:lim05hid16}, BC-Diffusion only learns the action distributions in the dataset, excluding the state information in OOD states. As an ablation study, we provide illustrations with different scaling parameters that control the weight of state reconstruction loss in Fig. \ref{fig:srdpabl08}. We use a dual head architecture with hidden layer sizes $(16^{shared}, 16^{shared}, 16)$ for SRDP and three hidden layers with size 16 for BC-Diffusion. Notice that SRDP with scaling parameter 0 performs similarly to the BC-Diffusion baseline in Fig. \ref{fig:srdpabl08} as this reduces SRDP to BC-Diffusion. We compute Chamfer distances between the ground truth actions and the actions generated by SRDP with varying scaling parameters ($\lambda$) and BC-Diffusion for each group of states. Table \ref{table:abl0chamfer}, shows the sum of Chamfer distances over each group of states across five random seeds, at 4000 intervals. Chamfer distance is particularly suitable for our case because it measures the distance between two sets of points. BC-Diffusion is prone to overfitting since the chamfer distance increases as the training progresses. The results show that SRDP performs significantly better and converges faster than BC-Diffusion with appropriate scaling parameter selection.

\begin{table}[!t]
\caption{Ablation Study for SRDP Scaling Parameters over training iterations at 4000 intervals with $\mathbf{Chamfer}$ $\mathbf{Distance}$.\label{tab:table1}}
\centering
\scalebox{0.7}{
\begin{tabular}{l||c|c|c|c|c}
\hline  \hline Method &4000 &8000  &12000 &16000 &20000\\
\hline $SRDP(\lambda=0.0)$  & $1.62\pm0.31$& $1.51\pm0.06$& $1.63\pm0.26$&$1.61\pm0.34$& $1.63\pm 0.32$ \\
\hline $SRDP(\lambda=0.25)$ &$1.49\pm0.25$ &$1.03\pm0.59$ &$0.91\pm0.62$ & $0.82\pm0.64$&$0.81\pm 0.67$ \\
\hline $SRDP(\lambda=0.5)$  &$\boldsymbol{1.1}\pm0.3$ & $1.07\pm0.55$&$1.01\pm0.72$ & $0.91\pm0.68$&$0.79\pm 0.55$ \\
\hline $\mathbf{SRDP(\lambda=0.75)}$ & $1.62\pm0.24$& $\boldsymbol{0.88}\pm0.51$&$0.83\pm0.43$ & $0.54\pm0.28$&$\boldsymbol{0.39}\pm 0.16$\\
\hline $SRDP(\lambda=1.0)$ &$1.24\pm0.48$ &$0.95\pm0.35$ & $\boldsymbol{0.68}\pm0.55$& $\boldsymbol{0.47}\pm0.43$&$0.44\pm 0.41$ \\
\hline\hline $BC$-$Diffusion$ &$1.41\pm0.01$&$1.51\pm0.16$&$1.54\pm0.18$&$1.6\pm0.14$&$1.62\pm0.15$\\
\hline
\end{tabular}}
\label{table:abl0chamfer}
 \vspace*{-1.5mm}
\end{table}

\subsection{Multimodal UR10 Experiments}
\begin{figure}[htbp]
 \vspace*{-1.5mm}
\includegraphics[width=\columnwidth]{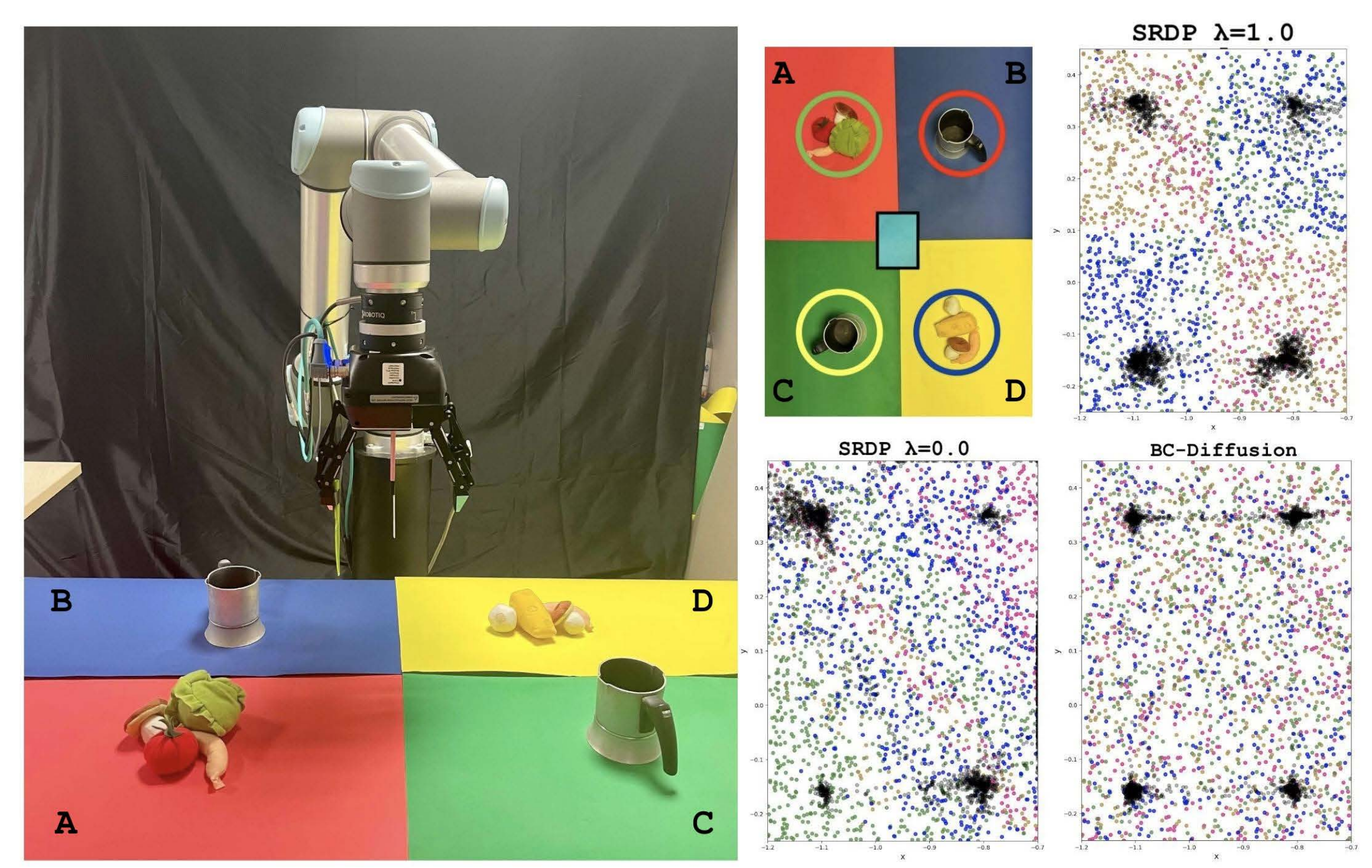}
 \caption{Real robot setup and the comparison of SRDP$(\lambda=1.0)$ with BC-Diffusion and $SRDP(\lambda=0.0)$. State samples in the training dataset are within the light blue region. The color-coded regions are labeled as follows: A (red/magenta), B (blue), C (green), and D (yellow/light brown).}
	\label{fig:ur10figs}
  \vspace*{-2mm}
\end{figure}

We design a multimodal UR10 robot experiment to illustrate the application of our method in the real world and highlight the importance of OOD generalization. Our hardware setup consists of a 6-Dof UR10 manipulator robot mounted with a Robotiq gripper illustrated in Fig. \ref{fig:ur10figs}. Similar to the experiments in \cite{chi_diffusion_visiomotor} designed for a visuomotor policy learning task, we define the actions for our policy as the space positional commands of the end-effector. In the setting visualized in Fig. \ref{fig:ur10figs}, the robot learns to reach the vegetables or the coffee maker from a random initial state. If the gripper is opened in states in region A (red/magenta) or D (yellow/light brown) (enabling the use of the pink coffee stirrer), it should reach the coffee maker for stirring. Conversely, if the gripper is enclosed (allowing the use of the green knife) in region B (blue) or C (green), it should reach the vegetables for chopping. The training dataset is constructed from the true space positional commands of the end-effector, $s_{train}\sim \mathcal{U}([-1.0,-0.9]\times[0.03,0.17]$. We sample states from the uniform distribution $\mathcal{U}([-1.2,-0.7]\times[-0.25,0.45])$ to evaluate the policies in OOD states. The action distributions are generated with the same procedure outlined in Eq. \ref{eq:actiongen} though the origin and GMM distributions are shifted for real UR10 space. Specifically, for states with respect to the center position at $(-0.95, 0.1)$, action distributions for the right upper, left upper, left bottom, and right bottom regions are produced with corresponding mean $\mu_1(s)=(-0.8,0.35)$, $\mu_2(s)=(-1.1,0.35)$, $\mu_3(s)=(-1.1,-0.15)$, $\mu_4(s)=(-0.8,-0.15)$ and diagonal covariance matrix $\boldsymbol{\Sigma}$ with $\sigma=0.015$. We evaluate SRDP and baselines trained with 75 diffusion timesteps across five random seeds. Chamfer distance results of $\mathbf{SRDP}(\lambda=1.0)$  compared to the baseline with respect to the ground truth contextual action distribution in the dataset is $(\boldsymbol{0.071}\pm 0.02)$. For Diffusion-QL and SRDP$(\lambda=0.0)$ the chamfer distances are $(0.27\pm 0.01)$, $(0.28\pm 0.02)$ respectively. We use a dual head architecture with hidden layer sizes $(32^{shared}, 16^{shared}, 32)$ for SRDP$(\lambda=0.0)$ and SRDP$(\lambda=1.0)$ and $(16,16,16)$ for BC-Diffusion. SRDP$(\lambda=0.0)$ is BC-Diffusion with a bottleneck layer in the middle without a state reconstruction loss. The action generation time for SRDP across 10 trials is $8.34\pm0.28$ ms on an NVIDIA 4090 GPU. Results indicated that the architecture change did not imply a better OOD generalization; hence state reconstruction signal is essential in learning multimodal expert behavior from OOD data.
\subsection{Missing Data Maze Environment}
\begin{figure}[htbp]
 \vspace*{1mm}
	\centering
         \subfigure[]{
		\includegraphics[width=0.185\columnwidth]{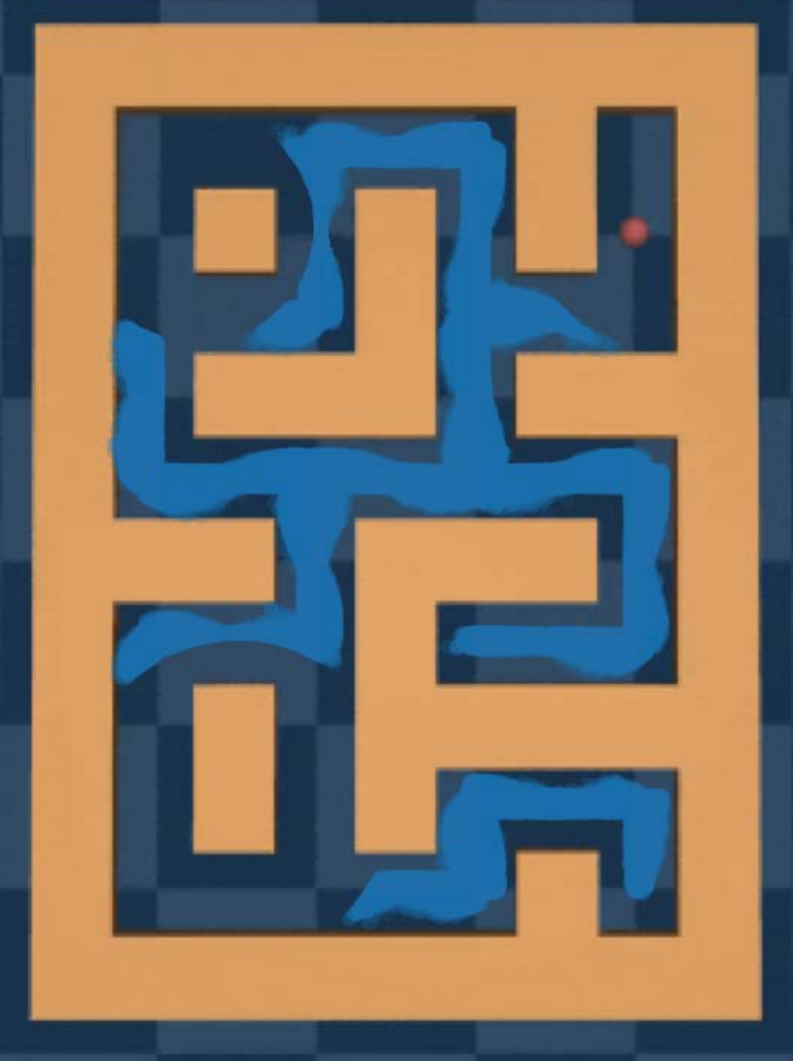}}
       \subfigure[]{
		\includegraphics[width=0.19\columnwidth]{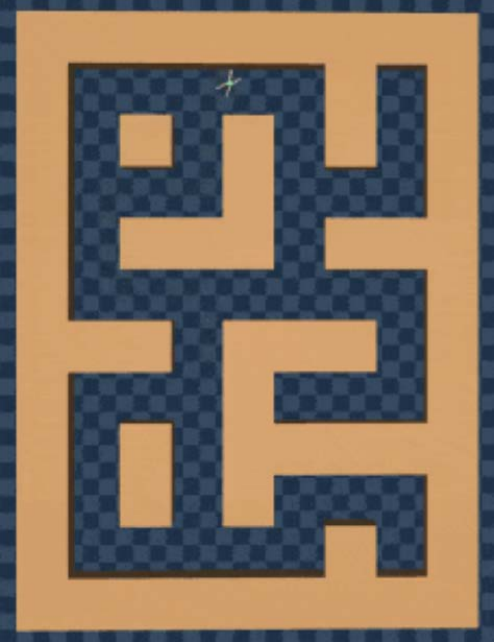}}
            \subfigure[]{
	\includegraphics[width=0.25\columnwidth]{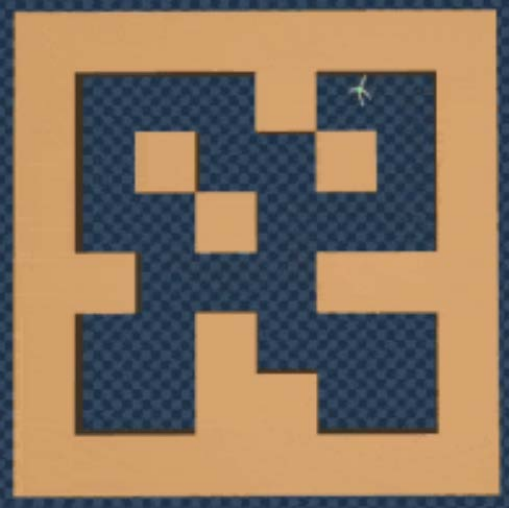}}
        \subfigure[]{
		\includegraphics[width=0.25\columnwidth]{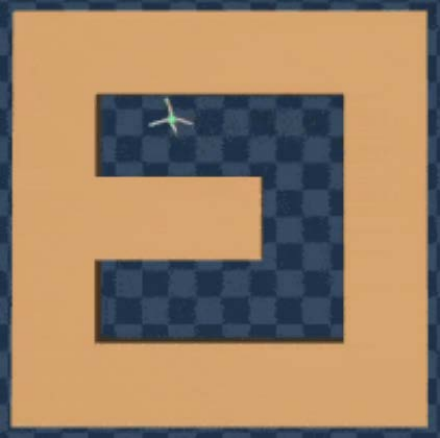}}
        \caption{(a) Maze2d environment with missing data ``maze2d-missing-data-large-v1" from \cite{mark2023offline}. D4RL AntMaze Environments with three layouts: (b) Antmaze-large-v0, (c) Antmaze-medium-v0, (d) Antmaze-umaze-v0.}
	\label{fig:antmazed4rl}
        \subfigure[Half-cheetah]{
		\includegraphics[width=0.30\columnwidth]{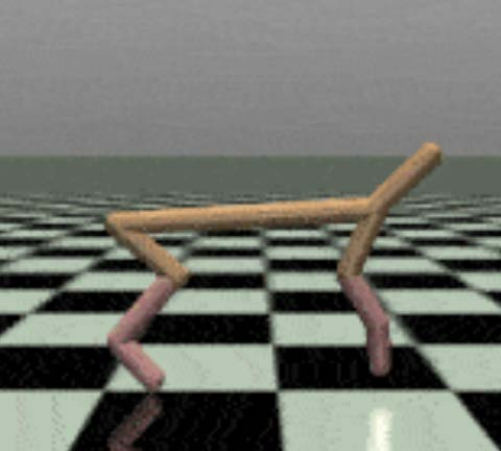}}
        \subfigure[Hopper]{
		\includegraphics[width=0.27\columnwidth]{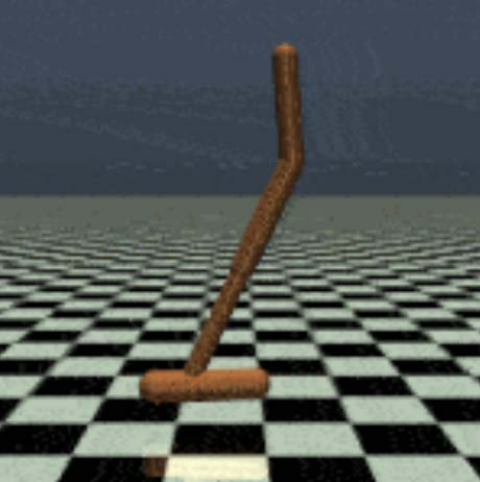}}
        \subfigure[Walker2d]{
		\includegraphics[width=0.30\columnwidth]{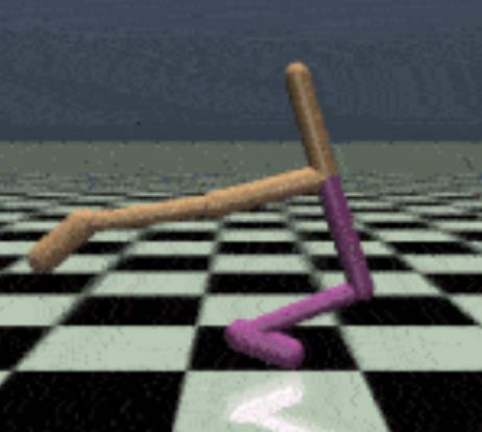}}
	\caption{D4RL Gym-MuJoCo Environments}
	\label{fig:gymd4rl}
\end{figure}

Maze2d navigation environments were introduced as benchmarks for offline RL in D4RL \cite{fu2020d4rl}. In this environment, a 2D force-actuated ball robot must navigate through a closed maze to reach a fixed goal location during evaluation. The maze2d dataset consists of trajectories collected by a planner agent, which uses a PD controller to reach a random goal from a random initial state location. The actions are defined as the linear force applied in x-y directions, whereas the observations are defined as the concatenation of the position and the linear velocity of the ball in the x-y direction.
\begin{table}[!t]
\caption{ Comparison of SRDP with Diffusion-QL baseline. \label{tab:missmazetable}}
\centering
\begin{tabular}{l||c}
\hline  \hline \text{maze2d-missing-data-large-v1}  & Normalized Score\\
\hline $\mathbf{SRDP(\lambda=1.0)}$ &$\boldsymbol{35.0}\pm 28.2$ \\
\hline $\mathbf{SRDP(\lambda=0.75)}$ &$23.9\pm 20.4$ \\
\hline $\mathbf{SRDP(\lambda=0.25)}$& $20.0\pm37.9$ \\
\hline $\mathbf{SRDP(\lambda=0.0)}$ &$1.7\pm2.3$ \\
\hline  $\mathbf{Diffusion-QL}$ & $13.1\pm 15.3$ \\
\hline \hline \text{maze2d-large-v1}  & Normalized Score\\
\hline $\mathbf{SRDP(\lambda=0.75)}$&$\boldsymbol{203.6}\pm 19.7$\\
\hline $\mathbf{Diffusion-QL}$ & $ 189.1 \pm 15.3$ \\
\hline
\end{tabular}
\end{table}
To assess the performance of SRDP in OOD states, we use the sparse reward maze2D offline RL dataset ``maze2d-missing-data-large-v1" by \cite{mark2023offline} illustrated in Fig. \ref{fig:antmazed4rl}(a) where the data collected in the vicinity of three circles, one encapsulating the goal, is missing. This task is used to evaluate the performance of online and offline finetuning after offline pretraining in \cite{mark2023offline}. In our context, we focused on results without online finetuning, aligning with the more challenging experimental framework adopted throughout this study. Notably, the agent starts from a random initial state during evaluation, and a significant portion of state data is absent near the goal and random initial state locations. An ablation study on this environment demonstrating the advantage of SRDP in Table II shows the mean and standard deviation of normalized scores obtained across five random seeds. More specifically, the results indicate the superiority of $\mathbf{SRDP(\lambda=1.0)}$ compared to Diffusion-QL \cite{wang2023diffusion} with $(256, 256, 256)$. It is important to note that SRDP with a dual head architecture with hidden layer sizes  $(512^{shared}, 256^{shared}, 512)$ and scaling parameter $\lambda=0.0$ becomes Diffusion-QL without a state reconstruction decoder. In addition, SRDP performs superior to the baseline for ``maze2d-large-v1" sparse environment in D4RL without missing data.

\subsection{D4RL Benchmark}
D4RL benchmark \cite{1606.01540}, comprising extensive robotics datasets, has been widely used as a benchmark for offline RL algorithms. In AntMaze environments, the objective of an 8-DoF quadruped ant robot is to navigate to a 2D goal location in various mazes. Correspondingly, in Gym-MuJoCo (\cite{1606.01540,todorov2012mujoco}) half-cheetah, hopper, and walker2d environments, the objective is to achieve forward locomotion. Since offline RL differs from imitation learning, each dataset in Table \ref{tab:table3} includes various amounts of suboptimal data. For these experiments, we follow the details of the online implementation used in Diffusion-QL, where they assume that the policies can be evaluated at fixed intervals with few online interactions. This is a form of early stopping in RL introduced in \cite{ada_ugur_akin_2022} where the hyperparameter, the number of training epochs, is tuned for Diffusion-QL and the proposed model, SRDP. We report the average normalized scores of undiscounted returns for SRDP and Diffusion-QL in Table \ref{tab:table3}, where 100 corresponds to expert-level behavior compared to the normalized scores for Diffusion-QL from \cite{wang2023diffusion}.

\begin{table}[!t]
\caption{Comparison of SRDP with the Diffusion-QL baseline. \label{tab:table3}}
\centering
\begin{tabular}{l||c|c}
\hline  \hline $\mathbf{AntMaze}$  & $\mathbf{Diffusion}$-$\mathbf{QL}$&$\mathbf{SRDP(ours)}$ \\
\hline \text{antmaze-umaze-v0}  & 96.0& $\boldsymbol{96.4}\pm1.5$\\
\hline \text{antmaze-umaze-diverse-v0}  & 84.0&$\boldsymbol{89.8}\pm 5.1$\\
\hline \text{antmaze-medium-play-v0}  & $\boldsymbol{79.8}$ &$78.4\pm8.1$\\
\hline \text{antmaze-medium-diverse-v0} & 82.0 &$\boldsymbol{90.6}\pm6.0$\\
\hline \text{antmaze-large-play-v0} & 49.0&$\boldsymbol{63.0}\pm 8.0$\\
\hline \text{antmaze-large-diverse-v0}   & 61.7&$\boldsymbol{62.6}\pm 3.5$\\
\hline $\mathbf{Average}$ & 75.4& $\boldsymbol{80.1}$ \\
\hline \hline $\mathbf{Gym}$ & $\mathbf{Diffusion}$-$\mathbf{QL}$&$\mathbf{SRDP(ours)}$\\
\hline \text{halfcheetah-medium-v2}& 51.5 &$\boldsymbol{51.9}\pm 1.5$\\
\hline \text{hopper-medium-v2} & 96.6 &$\boldsymbol{96.8}\pm 3.1$\\
\hline \text{walker2d-medium-v2} & 87.3&$\boldsymbol{88.1}\pm0.5$\\
\hline \text{halfcheetah-medium-replay-v2} & $\boldsymbol{48.3}$ &$48.1\pm0.3$\\
\hline \text{hopper-medium-replay-v2} & 102.0 &$\boldsymbol{102.1}\pm0.1$\\
\hline \text{walker2d-medium-replay-v2} & 98.0&$\boldsymbol{98.1}\pm 1.2$\\
\hline \text{halfcheetah-medium-expert-v2} & 97.2 &$\boldsymbol{97.5}\pm 0.1$\\
\hline \text{hopper-medium-expert-v2} & 112.3 &$\boldsymbol{112.6}\pm 0.4$\\
\hline \text{walker2d-medium-expert-v2} & $\boldsymbol{111.2}$ &$111.1\pm0.4$\\
\hline $\mathbf{Average}$ & 89.3 & $\boldsymbol{89.6}$\\
\hline
\end{tabular}
\end{table}

Antmaze datasets are generated following non-Markovian and suboptimal policies, sparse rewards, and multitask data design procedures \cite{fu2020d4rl}. Fig. \ref{fig:antmazed4rl} (b), (c), (d) shows the AntMaze environments where each maze layout, large, medium, umaze, has different levels, such as play and diverse. The performance scores across five random seeds in Table \ref{tab:table3} demonstrate that the proposed SRDP model is superior for all cases except ``antmaze-medium-play-v0''. OOD generalization is beneficial in multitask settings; hence, SRDP can enhance performance significantly.

Gym-MuJoCo environments, comprising continuous control tasks in MuJoCo \cite{todorov2012mujoco}, have been commonly used in deep RL \cite{fu2020d4rl,chen2021decisiontransformer,fujimoto2021a,haarnoja2018soft,brandfonbrener2021offline}. In D4RL, datasets are collected using the online RL interaction data of a Soft Actor-Critic (SAC) agent \cite{haarnoja2018soft, fu2020d4rl}. To evaluate an offline-RL algorithm in narrow data distribution and suboptimal behavior policy settings, ``medium", ``medium-replay", and ``medium-expert" datasets are generated. Medium datasets are generated by collecting the rollouts from a medium-performing policy, whereas medium-replay datasets are generated by keeping a replay buffer of rollouts until the RL policy reaches a medium-level performance. Medium-expert datasets include expert trajectories by 50\% in addition to medium-level rollouts. Results presented in Table \ref{tab:table3} show that our proposed model, SRDP, performs better than Diffusion-QL \cite{wang2023diffusion} across five random seeds.

\section{Conclusion}
In this work, we propose SRDP for OOD generalization in offline RL, a representation learning-based method built on top of the recent class of diffusion policies introduced by \cite{wang2023diffusion}. To show the multiple skills learned by the model when evaluated in OOD states, we designed a 2D Multimodal Contextual Bandit environment and applied it to a real robotic scenario. Compared to prior work, our results indicate that SRDP can represent multimodal policies, partition the state space more accurately, and converge faster in real-robot and simulation environments. In addition, results show that SRDP can learn superior models compared to previous work in Antmaze and Gym-MuJoCo environments in D4RL benchmarks \cite{fu2020d4rl} with various levels and agents. Finally, on a sparse continuous control navigation task where critical regions of the state space are completely removed from the offline RL dataset, our method performs significantly better than the standard diffusion policy-based RL. For future work, we plan to integrate a vision module into SRDP, reduce the inference time in diffusion policies, and extend our approach to trajectory-level diffusion probabilistic models \cite{janner2022diffuser}.

\appendix
\begin{figure}[htbp]
	\centering
  \includegraphics[width=\columnwidth]{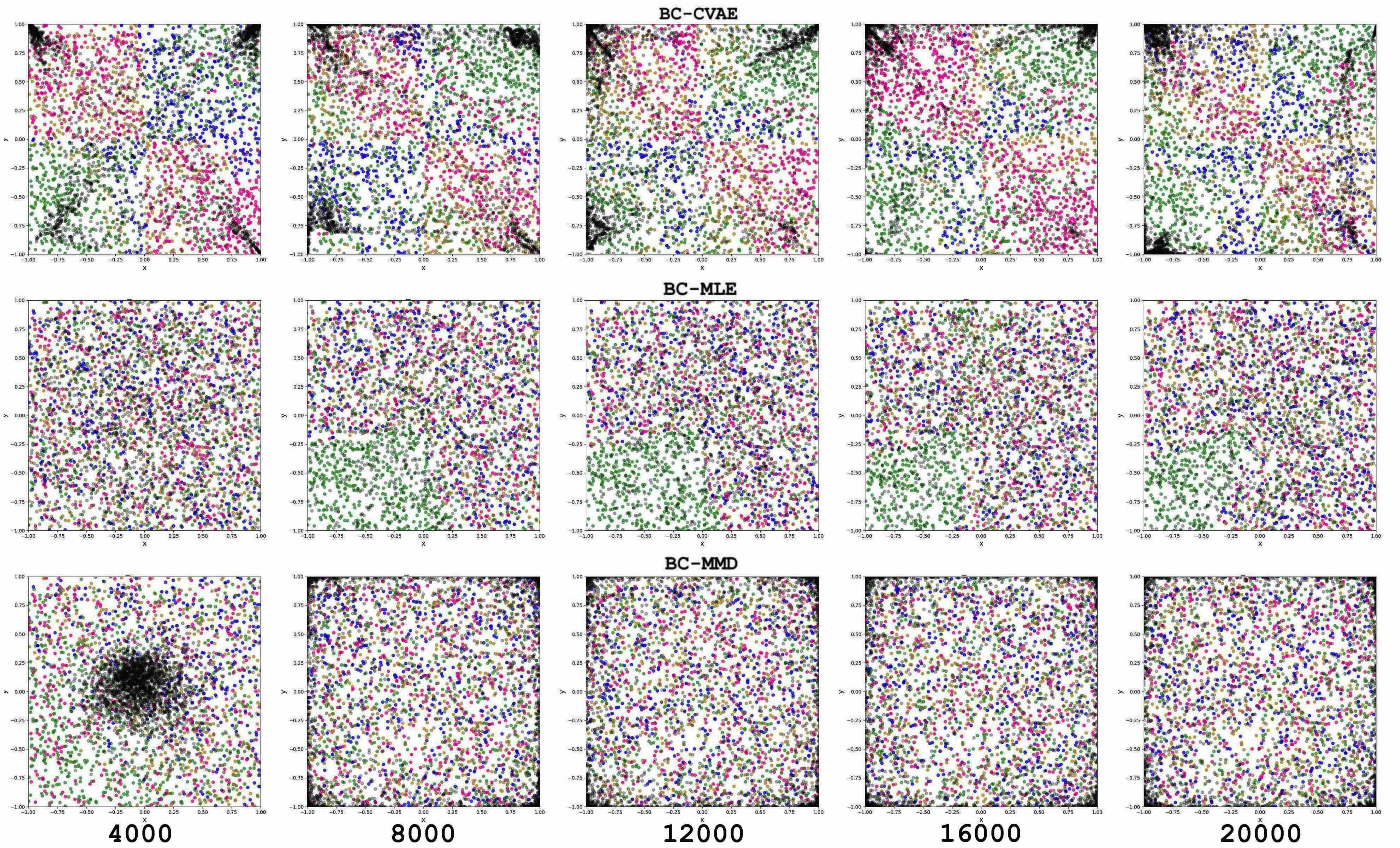}
 \caption{Additional results for Section V-A. Top to bottom Row: BC-CVAE represents the policy by a conditional variational autoencoder (CVAE) \cite{Fujimoto2018OffPolicyDR}, BC-MLE\cite{fujimoto2021a} uses a Multivariate Gaussian to model the policy with a diagonal covariance matrix, BC-MMD \cite{DBLP:journals/corr/abs-1906-00949} uses a CVAE that learns the behavior policy to constrain a Gaussian policy. The training dataset is constructed from $s_{train} \sim \mathcal{U}([-0.08,0.08]\times[-0.08,0.08])$. The number of training iterations increases incrementally by 4000 from left to right.}
\label{fig:lim08hid16}
\end{figure}

\section*{Acknowledgment}
This research was supported by Grant-in-Aid for Scientific Research – the project with number 22H03670, the project JPNP16007 commissioned by the New Energy and Industrial Technology Development Organization (NEDO), the Scientific and Technological Research Council of Turkey (TUBITAK, 118E923), and INVERSE project (101136067) funded by the European Union. The numerical calculations reported in this paper were partially performed at TUBITAK ULAKBIM, High Performance and Grid Computing Center (TRUBA resources). The authors would like to thank the anonymous reviewers and Editor Aleksandra Faust for their valuable feedback and constructive suggestions.
%%%%%%%%%%%%%%%%%%%%%%%%%%%%%%%%%%%%%%%%%%%%%%%%%%%%%%%%%%%%%%%%%%%%%%%%%%%%%%%%

\bibliographystyle{IEEEtran}  
\bibliography{IEEEabrv,references}
\end{document}